\documentclass{article}

\usepackage[nonatbib,final]{neurips_2023}

\usepackage{caption}
\usepackage{makecell}
\usepackage{adjustbox}

\usepackage[utf8]{inputenc} 
\usepackage[T1]{fontenc}    
\usepackage{hyperref}       
\usepackage{url}            
\usepackage{booktabs}       
\usepackage{amsfonts}       
\usepackage{nicefrac}       
\usepackage{microtype}      
\usepackage{xcolor}         
\usepackage{booktabs}
\usepackage{amsmath}
\usepackage{placeins}
\usepackage{subcaption}

\usepackage{biblatex}

\title{Krutrim LLM: Multilingual Foundational Model for over a Billion People}

\author{%
 Aditya Kallappa, Palash Kamble, Abhinav Ravi, Akshat Patidar, Vinayak Dhruv, \\ 
 \textbf{Deepak Kumar, Raghav Awasthi, Arveti Manjunath, Himanshu Gupta, Shubham Agarwal,} \\
 \textbf{Kumar Ashish, Gautam Bhargava, Chandra Khatri}
 \thanks{Additional Contributors:
Sanket Shah, Sulabh Katiyar, Sindhu Pawar, Soham Pendurkar, Pranav Raveendran, Bidyapathi Ray, Daud Ibrahim, Divyansh Rajput, Pidathala Sowjanya, Rahul Kumar, Rishabh Nahata, Pranav Raveendran, Bidyapathi Ray, Prateek Shrivastava, Yogendra Mishra, Azhagiri S, Priyanka Nayak, Sandesh Bafna, Aniruddha Uttam Tammewar, Vivek Dahiya, Ali Faraz, Ayush Tarun, Shaharukh Khan, Shubham Kakde, Nishant kumar, Debanjana Biswas, Ashish Anand Kulkarni, Rajkiran Panuganti, Hareesh Kumar \\ \\
Acknowledgement: Ravi Jain, Bhavish Aggarwal
} \\
Krutrim AI Team, Bangalore, India}

\begin{document}

\maketitle

\begin{abstract}
  India is one of the most vibrant and culturally diverse societies. Developing a general-purpose artificial intelligence system tailored for the Indian market presents unique challenges. These include accounting for the nation's cultural nuances, accommodating its linguistic diversity with numerous regional languages, adapting to the prominence of oral traditions, ensuring accessibility to relevant data sets, and achieving scalability to serve the vast population effectively. Careful consideration and innovative approaches are necessary to navigate these complexities successfully. Existing foundation models for natural language tasks are predominantly trained on English data, limiting their effectiveness for languages native to India's over 1 billion citizens. Thousands of regional languages, dialects, language or code mixing pose representation challenges, exacerbated by sparse training data; Indic languages comprise just 1\% of Common Crawl corpora despite India representing 18\% of the global population. Consequently, lack of Indic language relevance and context representation leads current models to exhibit cultural and linguistic biases oriented towards Western contexts. We present Krutrim Large Language Model (LLM), a 2 trillion token multilingual foundation model designed to serve Indian demographic needs through equitable representation of the country's array of native tongues. Training data incorporates the largest known Indic language dataset, mitigating associated data scarcity obstacles that encumber model parity across dialects. Evaluations demonstrate Krutrim’s strong performance on Indic language benchmarks, surpassing or at par with state-of-the-art models despite being significantly smaller in training flops. Krutrim LLM also matches or exceeds standards set on English benchmarks by models trained on comparable flops (e.g. vs LLAMA-2 on 10 out of 16 tasks with average score of 0.57 vs 0.55 of LLAMA-2), evidencing flexible multilingual fluency. We further integrated search to deliver real-time and more factually relevant information via Krutrim LLM conversational app, working to make next-generation AI widely accessible for a diverse set of over 1 billion worldwide users. Through intentional design choices that redress endemic data imbalances, Krutrim LLM signifies meaningful progress in the pursuit of ethical, globally representative AI foundation models.
\end{abstract}

\section{Introduction}

\subsection{Challenges in Developing an India-Centric Large Language Model (LLM)}

Building Artificial General Intelligence (AGI) and AI models to cater to the diverse cultural context of India presents multifaceted challenges. The country's linguistic diversity\cite{Languages_of_India}, with hundreds of languages and dialects, poses a significant hurdle \cite{mhrd}. These languages span 4 major language families with Indo-Aryan and Dravidian being the most prominent ones~\cite{kakwani2020indicnlpsuite}. Many of these languages thrive in oral traditions, leading to frequent mixing and evolving linguistic patterns that are not well-documented digitally. This oral culture complicates the collection and digitization of data necessary for training robust AI systems. Additionally, India's rich tapestry of social and economic backgrounds adds another layer of complexity. The vast differences in socio-economic statuses influence digital access and technology usage, making it challenging to develop AI solutions that are inclusive and equitable. Furthermore, cultural nuances, which vary widely across regions, need to be understood and respected in AI applications to ensure they are relevant and sensitive to users' needs. The combination of linguistic diversity, cultural richness, and socio-economic disparities requires innovative approaches to data collection, model training, and the development of algorithms that can adapt to and reflect India's multifaceted society.

Most recent AI models, like, LLaMA \cite{touvron2023llama}, GPT-3.5, and others, face significant limitations when representing the Indian ethos and languages due to the unique linguistic and cultural challenges prevalent in the region \cite{brown2020language}. 
In fact, the structure of some Indian languages, such as Sanskrit, which allows for the creation of virtually infinite compound words, presents a stark contrast to English's finite vocabulary. These languages are not only syntactically complex but are also semantically rich, making them difficult to model with the tokenizers designed primarily for Western languages. Such systems often result in inefficient processing of Indian languages, leading to excessively long sequences that can hamper the effectiveness of AI models.

Additionally, the reliance on digitized data for training these models poses a challenge, as there is a significant lack of local, undigitized data reflecting the vast cultural and linguistic diversity of India. This scarcity of data is compounded by the fact that Indian languages are under-represented in major digital corpora. For instance, Indian languages account for a mere 1\% of the Common Crawl\footnote{\url{http://commoncrawl.org/}} dataset~\cite{buck2014n,penedo2023refinedweb}, despite India constituting 18\% of the global population. This under-representation leads to biased models that do not adequately capture the nuances of Indian languages and culture. The poor representation of Indic data not only affects the accuracy and relevance of these AI models for Indian users but also highlights a broader issue of inclusivity and representation in the development of global AI models.

\subsection{Contributions}
The development of Krutrim LLM, India's first and premier Large Language Model (LLM), represents a monumental stride towards creating AI technologies that are deeply aligned with the country's linguistic and cultural diversity. The project embarked on an ambitious journey by curating an extensive range of Indic data from across the web, encompassing the myriad languages spoken throughout India. This foundational work enabled the training of the LLM on over 2 trillion tokens, a scale unprecedented in the context of Indian language processing. Recognizing the unique challenges of Indian languages, the team developed a specialized Indic tokenizer, tailored to efficiently process the complex morphologies and syntaxes inherent to these languages.

To enhance the model's performance further, Krutrim LLM incorporates state-of-the-art attention mechanisms such as Grouped Query Attention (GQA) \cite{ainslie2023gqa} and AliBi \cite{press2022train}, which significantly improve its ability to handle longer contexts and provide faster responses, thereby elevating the user experience. This advanced technical framework allows Krutrim LLM to outperform several open-source models of comparable size or computational requirements, particularly in its handling of Indic languages, without any compromise.

Furthermore, the model underwent India-centric fine-tuning, covering a broad spectrum of topics and tasks relevant to the Indian context. This fine-tuning ensured that the model could effectively engage with and address the specific needs and nuances of Indian users. A linguistically and socially rich alignment was achieved through Direct Preference Optimization (DPO) \cite{rafailov2023direct}, ensuring that Krutrim LLM adheres closely to the values and preferences of its intended audience. This meticulous approach to development reflects a deep commitment to inclusivity and representation, setting a new standard for AI models designed to serve the rich tapestry of India's languages and cultures.

Lastly, to make Krutrim LLM more functional for general purpose use cases and to make it factually accurate, we integrated web search. Krutrim LLM is accessible for general audience via our conversational app: \url{https://chat.olakrutrim.com/} 

\section{Related Work and Background}
There have been rapid advancements in the field of AI since the Transformers \cite{vaswani2023attention} came out. It has led to notable increase in endeavors dedicated to constructing versatile conversational AI systems adept at comprehending and engaging with users across multiple topics and open ended conversations. 

General purpose LLMs are fundamentally changing how software and applications are being built which is further dramatically changing user experiences. Some of the most popular LLMs which have lead to this change are GPT-3.5 \cite{achiam2023gpt, brown2020language}, LLaMA \cite{touvron2023llama}, Gemini \cite{geminiteam2023gemini}, Anthropic Claude \cite{claude}, Mistral \cite{jiang2023mistral}, Inflection \cite{inflection}, and Grok-1 \cite{grok}. 

Brown et al.\cite{brown2020language} demonstrated the feasibility of generating coherent and contextually relevant responses in various languages, including Indic languages, using large language models.
Building upon the success of GPTs and Claude \cite{claude, Askell2021AGL, kadavath2022language}, subsequent research endeavors have further propelled the development of conversational AI in the Indic language domain. One notable participant in this progression is Gemini/Bard project \cite{geminiteam2023gemini}. It has significantly advanced the state-of-the-art in conversational AI by leveraging large-scale pre-training techniques. Gemini/Bard's approach has led to substantial improvements in language understanding and generation, particularly in low-resource languages such as Hindi, Bengali and Tamil. Mistral \cite{jiang2023mistral} addressed low performance in European languages by collecting and prioritizing European languages along with English.

As most of these LLMs are trained on English or high resource languages, there are several limitations in adapting them for regional or local use cases, both at knowledge level (pre-training) and inference level (high token to word ratio).  Performance of these state-of-the-art models on Indic languages is far from the performance on English and European languages and therefore remained an open problem. Furthermore, most of these LLMs are biased towards non-Indic regions due to significantly lower volume of Indic data and therefore do not work for countries like India that have a rich and diverse landscape, culture and languages. 

Several Indian-origin fine-tuned versions of LLMs have emerged, presenting a promising array of open-source models such as OpenHathi tuned Airavata \cite{gala2024airavata}, Gemma based Navarsa \cite{NavarasaTeluguLLMLabs}, Kannada LLaMA, Tamil LLaMA \cite{balachandran2023tamilllama}, Odia LLaMA \cite{kohli2023building}, and other vernacular models. These models are constructed upon the open source Llama2 architecture, leveraging cutting-edge techniques like Parameter Efficient Fine-tuning (PEFT) \cite{houlsby2019parameter,hu2021lora}. However, a predominant characteristic/limitations of these models is their focus on monolingual or bilingual generation capabilities and lack of Indic knowledge and sentence construction.

Furthermore, initiatives such as Perplexity AI \cite{perplexity}, BingChat \cite{bing} and You.com \cite{you} have also made noteworthy strides in advancing conversational AI by providing concise answers with source links for verification, making it noteworthy for more factual research and getting factual information by having robust Web search with Retrieval Augmented Generation (WebRAG) pipelines. These efforts emphasises the importance of AI assistants being more personalized, truthful, accurate, and transparent.

In the landscape of natural language processing models, particularly in the realm of Indian languages, there exists a noticeable gap in understanding the nuances of the Indian context, including local vernacular dialects, socio-economic structures, and cultural intricacies. Many existing models lack the depth required to effectively cater to these specific requirements. Thus, the necessity arises to develop indigenous systems tailored to the needs of Indian users.

In light of these advancements, the Krutrim LLM endeavors to enhance both accuracy and speed, positioning itself as a versatile and efficient general-purpose chat assistant. By synthesizing insights from existing Indian-origin LLMs and incorporating innovative methodologies, Krutrim aims to bridge the gap between state-of-the-art language models and the unique linguistic and cultural landscape of India.

In this report, we undertake an examination of current LLMs and frameworks, accentuating their attributes, capabilities, and constraints related to Indic use cases. We offer a comparative analysis, elucidating how Krutrim LLM, enhanced by WebRAG integration, stands in relation to these existing solutions.



\section{Data Collection and Tokenization}

\begin{figure}
\includegraphics[scale=0.4]{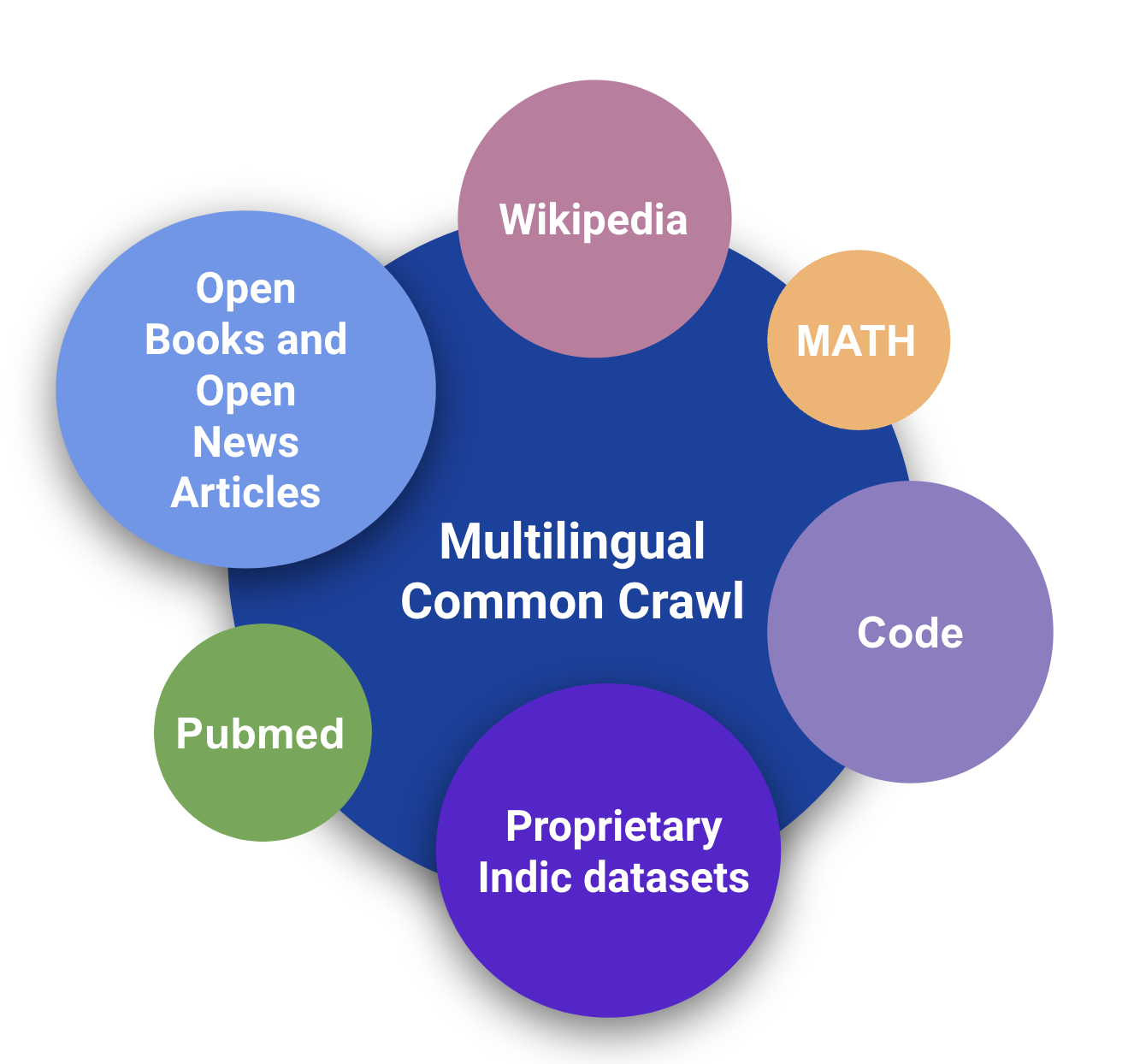}
\centering
\caption{Pre-training data sources.}
\label{Data_Sources}
\end{figure}

Krutrim LLM is trained on a comprehensive dataset encompassing a diverse range of sources such as open web and proprietary sources. Figure~\ref{Data_Sources} depicts various sources that were used in training Krutrim LLM. This dataset includes more than 2 trillion high-quality tokens meticulously crafted with a unique blend of data encompassing various languages, tasks, and sources. Notably, the dataset incorporates hundreds of billions of carefully curated Indic tokens, establishing Krutrim LLM as the largest known distribution of Indic data to date. We have trained new tokenizer based on sentencepiece\footnote{\url{https://github.com/google/sentencepiece}} byte pair encoding from scratch to represent EN and Indic languages.

\subsection{Data collection and cleaning}
For this study, a massive corpus comprising 2 trillion tokens was gathered through web scraping. The dataset, characterized by its unlabeled and unstructured nature, underwent essential cleaning processes, including de-duplication, removal of extra short passages and low quality text. In concurrent to our work, IndicLLMSuite \cite{khan2024indicllmsuite} depicts the techniques applied to collect high-quality Indic LLM training data. We made particular focus on Indic data sources like NDL, requiring additional cleaning and therefore more work. To achieve this, we followed similar techniques across diverse sources such as OpenWeb (open source subset of data from web archive) and leveraged open-source cleaned datasets like the RedPajama dataset subset, Books data, PubMed, Wiki, and StackFast. We applied techniques similar to Dolma \cite{soldaini2024dolma} and further enhanced the cleaning pipeline to address Indic languages.

\subsection{Data quality}
The data quality has momentous impact on the model's language learning and generation capability. Training with shorter prompts might result in generation of short responses and might also lead to non learning of a particular language in a multi lingual learning setting (e.g., observed in Hindi data). We also saw that providing large contexts only for a given language might not lead to learning of that languages altogether (e.g., observed in Gujarati data). The best setting for teaching a language is to show the model both smaller sentences and set of larger sentences. \\
Achieving a balanced distribution of pre-training data among different languages proves crucial for optimal language learning outcomes. We maintain substantial representation of each of the Indic languages.

\subsection{Tokenizer}
Existing open source tokenizers do not perform well on Indic languages leading to high token to word ratio. A sub-optimal tokenizer leads to inferior training and inference performance in terms of speed and accuracy. In order to address that, we trained the tokenizer from scratch optimized for English and Indic languages.

\section{Model and Architecture}
The Krutrim model architecture draws from the standard decoder only transformer framework \cite{vaswani2023attention}. 
We enlist key parameters of the model in table \ref{table:model_arch}. 
We train 7 billion parameter model on context length of 4096 tokens. We use ALiBi positional encoding method \cite{press2021train} which helps in expanding the context length. We also leverage GQA \cite{ainslie2023gqa} for faster inference and lower KV cache memory footprint. We use clipping of QKV matrix values for stable training. The standard ReLU activation function is used.

\begin{table}[]
\centering
\begin{tabular}{lllll}
\hline
Parameters                       & \multicolumn{1}{c}{Value} \\ \hline
Layers                 & 32                        \\ 
Number of KV heads     & 8                         \\ 
Number of attention heads & 48                     \\ 
Hidden dimension       & 4608                      \\ 
Sequence length        & 4096                      \\ \hline
\end{tabular}
\caption{Key parameters of the model.}
\label{table:model_arch}
\end{table}

\section{Training}

\subsection{Pre-training}
We pre-trained Krutrim LLM on a dataset of 2 Trillion tokens. The initial phase of unsupervised pre-training involves exposing the language model to vast datasets, allowing it to learn world knowledge and language capabilities through next word prediction. This critical stage sets the foundation for subsequent fine-tuning and specialized tasks.
In this stage of training, the model learns about world knowledge and language capabilities by the means of next word prediction. 

\subsubsection{Pre-training}
\begin{figure}
 \centering
 \includegraphics[scale=0.07]{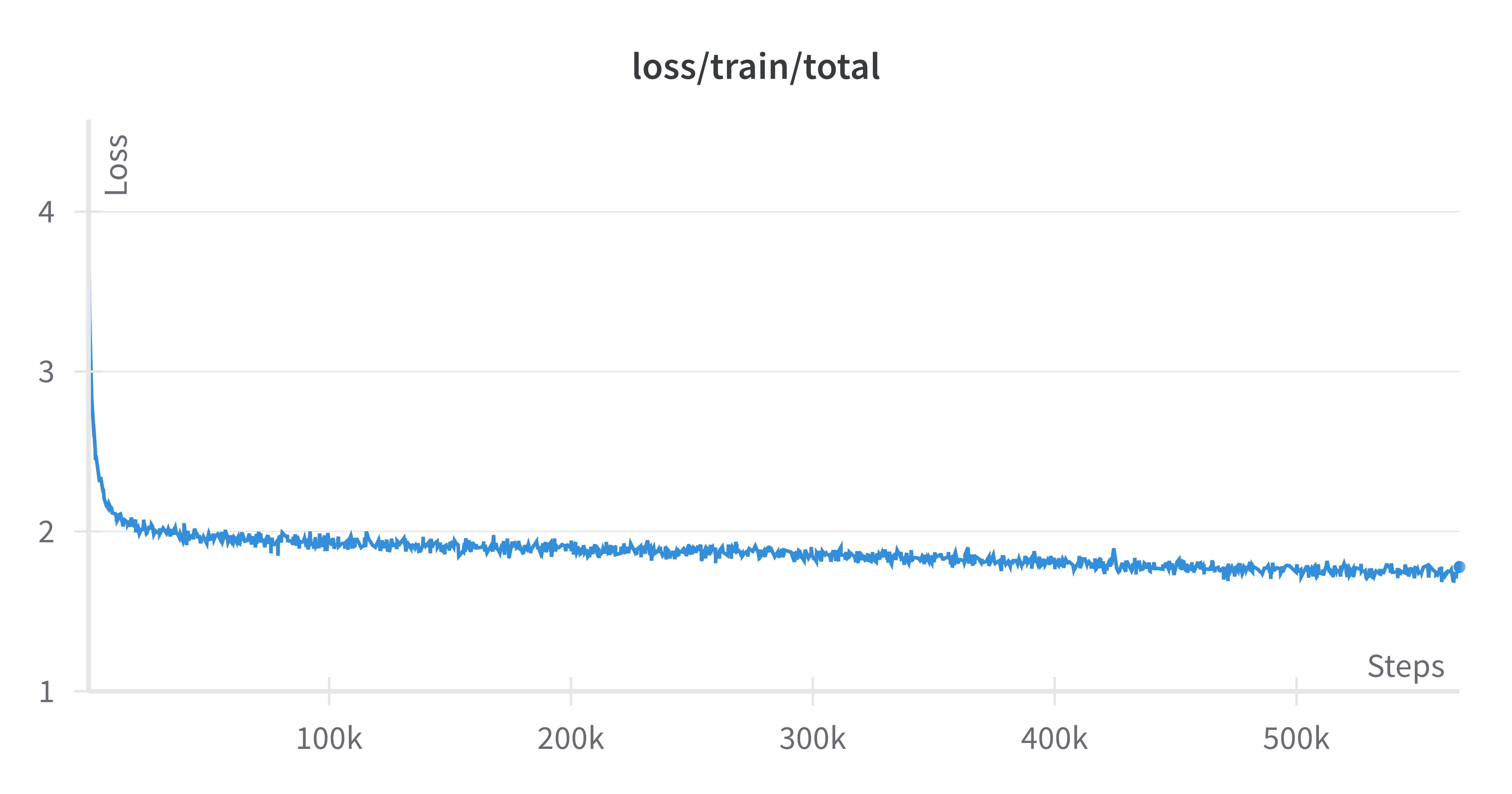}
 \caption{Pre-training loss}
 \label{fig:trainingloss}  
 \end{figure}

We trained Krutrim on H100 GPUs
resulting in $10^{23}$ FLOPS. To select the optimal model checkpoint post pre-training (PT), a systematic analysis of checkpoints at intervals of 20k checkpoints is conducted. Figure \ref{fig:trainingloss} depicts training loss during pre-training.

  \begin{figure}
 \centering
 \includegraphics[scale=0.23]{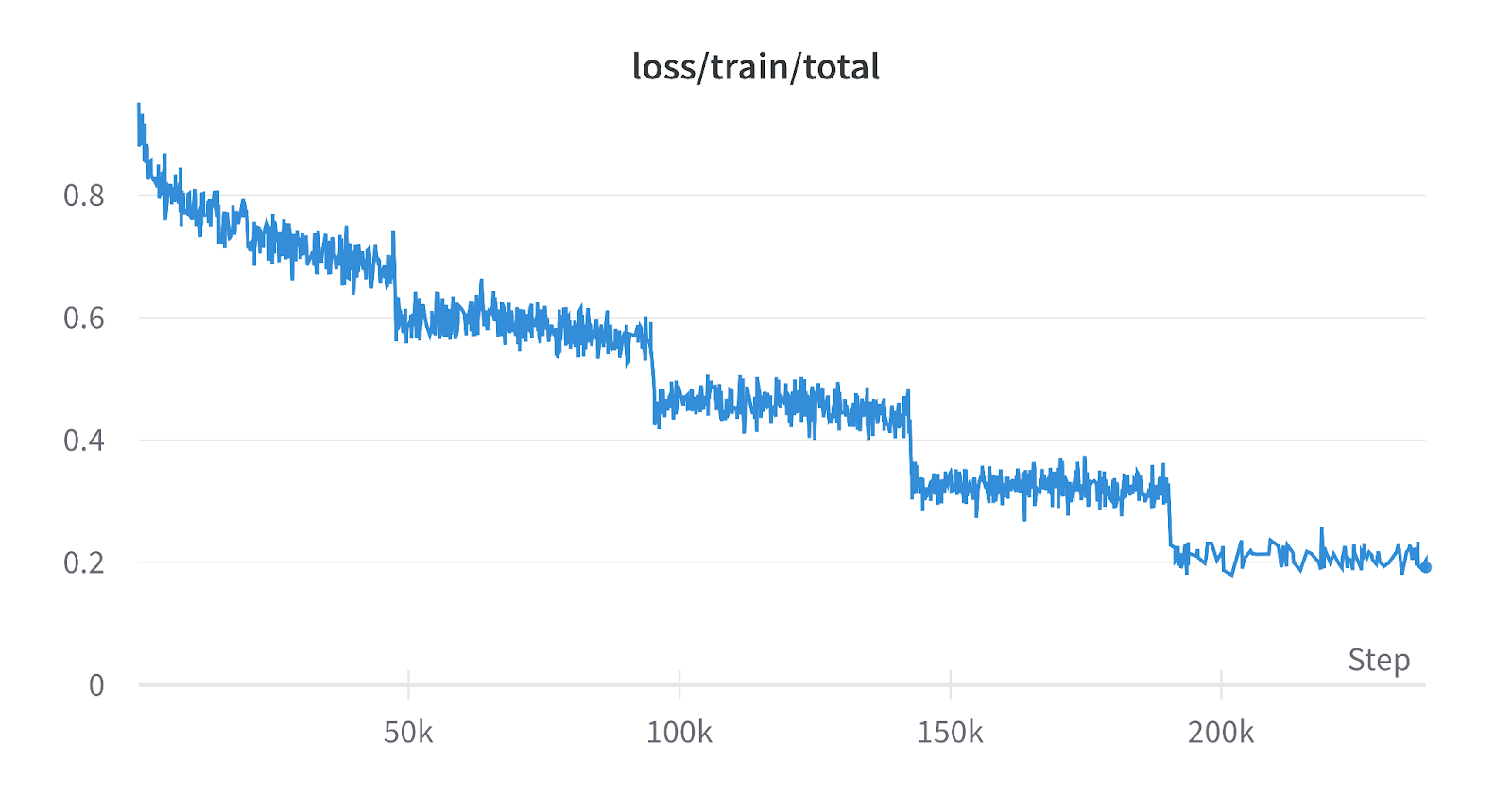}
 \caption{SFT loss}
 \label{fig:sfttrainingloss}  
 \end{figure}

\subsection{Continual Pre-training (CPT)} 
Continual Pre-training (CPT) is imperative in the realm of natural language processing, serving as a foundational element for adapting to multiple domains and acquiring diverse knowledge incrementally. This section delves into the necessity of CPT across various domains, highlighting its role in adapting language models to evolving requirements viz.

\begin{enumerate}

\item Facilitating the Acquisition of New Languages: CPT enables language models to swiftly learn new languages as needed. An illustrative example includes scaling from the current support for 10 languages to accommodating 22 Indic languages, showcasing the model's adaptability to linguistic diversity.

\item Staying Abreast of Fresh Knowledge and Domain-Specific Information: To effectively engage in specialized tasks, models must retain and apply domain-specific knowledge. CPT ensures models are well-versed in pertinent information, such as legal frameworks for legal argumentation or fundamental concepts in physics, chemistry, and mathematics to function as proficient tutors.

\item Enhancing and Expanding Model Capabilities: CPT serves as a mechanism for seamlessly incorporating new capabilities into existing models. This encompasses augmenting language models with programming skills like coding and code debugging, as well as reinforcing mathematical prowess to solve complex maths problems such as linear equations, LCM/HCF.

\end{enumerate}

In essence, CPT emerges as a pivotal approach for imbuing language models with the flexibility and adaptability necessary to navigate a spectrum of tasks, ranging from linguistic diversity and domain-specific expertise to the assimilation of new capabilities.

Though pretraining (PT) is important for imparting language learning and knowledge, CPT helps scenarios where we want to continuously add new language capabilities or domain knowledge without catastrophic forgetting of old knowledge and capabilities \cite{cossu2022continual}\cite{ke2023continual}. Unlike human learning, neural networks experience a significant drop in performance on previous tasks when learning new ones. We aim to minimise the catastrophic forgetting by employing a set of techniques discussed below.

\subsection{Instruction Tuning}
After the initial pre-training (PT) phase and in our case post CPT, the base language model acquired language generation capabilities and accumulates knowledge across Indic languages. However, it faces limitations in following instructions and engaging in conversations. The subsequent steps involve either supervised-Fine-Tuning (SFT) or Instruction-Fine-Tuning (IFT), each presenting a trade-off between knowledge retention and creativity learning. The challenge lies in finding the right balance, as infinite training on extensive datasets may result in forgetting of pre-training knowledge. Specifically, SFT introduces instruction-following capabilities but risks forgetting pre-training knowledge \cite{yang2024selfdistillation}.
\subsubsection{Knowledge Forgetting vs. Creativity Learning Trade-off}
The decision between SFT and IFT necessitates navigating the trade-off between knowledge retention and creativity learning. It is acknowledged that prolonged training on massive datasets may lead to a compromise in pre-training knowledge, necessitating strategic fine-tuning approaches.
\subsubsection{Instruction Tuning for Diverse Tasks}
We explored instruction tuning across various tasks to enhance the model's adaptability and performance. Tasks included:
\begin{itemize}
\item Translation (In-En and En-In): Teaching bidirectional translation capabilities.
\item Summarization: Enhancing the model's ability to generate concise summaries.
\item Chain of Thoughts (COT) Reasoning: Enhancing the model's sequential reasoning capabilities.
\item Single and Multi-turn Dialogues and Conversations: Teaching conversational engagement.
\item Safety around Sensitive Topics: Ensuring appropriate responses on sensitive subjects.
\item General Knowledge: Verifying the model's grasp of broad factual information.
\item Coding and Programming: Teaching coding-related tasks and instructions.
\item Chat Bot Self-Identification: Teaching the model about self-identity in a chat bot context.
\end{itemize}
\subsubsection{Task-Specific Personas}
Tasks were designed with diverse user personas in mind, including students, technical and non-technical workers, and creative content creators. The aim was to tailor the language model's performance to suit the varied needs and expectations of different user groups pertaining to India.


These comprehensive fine-tuning tasks contribute to shaping a language model that not only possesses general language capabilities but also excels in task-specific instructions, accommodating a range of user personas and real-world scenarios. The ongoing challenge lies in optimizing the trade-off between knowledge retention and task-specific learning during fine-tuning phases.
Figure \ref{fig:sfttrainingloss} depicts training loss during SFT.
\subsection{SFT for Answering Factual Questions}\label{webrag}
For factual questions, we observed that the base Instruction-tuned model hallucinated around 33\% of times giving false facts to the user. We also observed that for around 14\% of times the model hallucinated and fell prey to confirmation bias for adversarial and factual incorrect questions. To overcome this, we conduct an SFT over the base SFT model for teaching the model the following: (1) Answering factual questions strictly from the provided knowledge sources (2) Identifying any kind of ambiguity or factual incorrectness in the query and highlighting it instead of generating answer for the same (3) Refraining from answering those questions for which answer cannot be found in the provided knowledge sources.

\subsection{Alignment with Human Feedback}
Direct Preference Optimization (DPO) has shown superiority over Proximal Policy Optimization\cite{schulman2017proximal} (PPO)-based Reinforcement Learning from Human Feedback (RLHF)\cite{stiennon2022learning} in certain aspects. DPO, as highlighted in the study by Rafailov et al., has demonstrated better control over sentiment in text generation and comparable or improved response quality in summarising and dialogue tasks compared to PPO-based RLHF \cite{rafailov2023direct}. Additionally, experiments in \cite{rafailov2023direct} have indicated that DPO can align language models with human preferences effectively and with simplicity, eliminating the need for extensive hyperparameter tuning or complex sampling from the model during fine-tuning. In our experimentation, we employed a reduced learning rate of 5e-7 in conjunction with a $\beta$ value set to 0.1, leveraging a dataset comprising approximately 20,000 instances focused on safety topics. We scale data points in later phases for myriad of the alignment topics. Notably, our observations underscore the necessity of maintaining a balanced language data mixture to prevent the model from exhibiting forgetting behaviors related to other linguistic capabilities.

\section{Experimental Results and Analysis}

\subsection{Embeddings}
\begin{figure}
\includegraphics[scale=0.55]{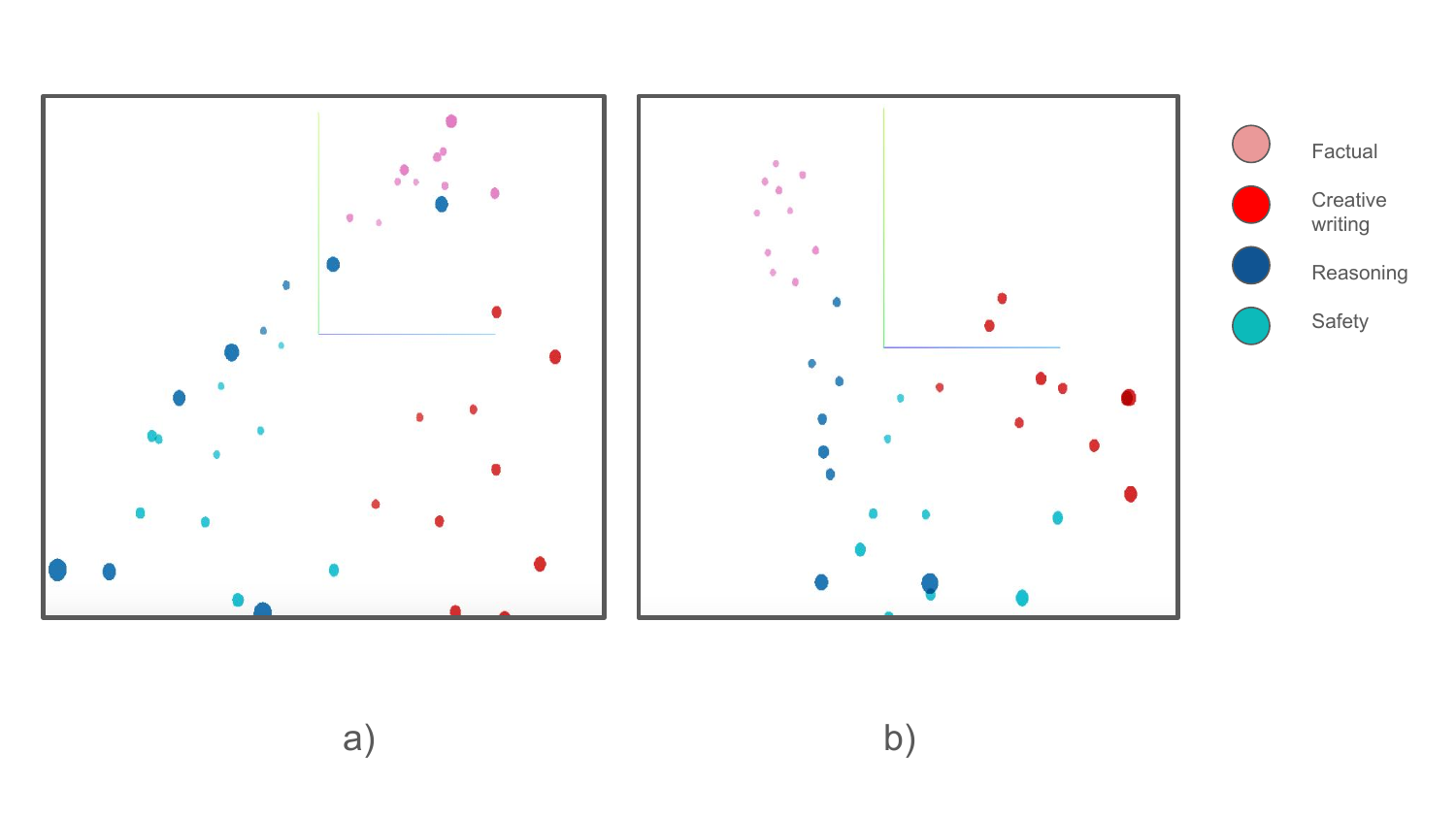}
\caption{The UMAP projection of embedding for randomly sampled tasks across 4 categories as shown in above figure. (a) The plot shows that embedding projection for the data points are intermixed and tasks are not separable after pre-training stage (b) The plot shows that Krutrim SFT model has better understanding of tasks and is able to separate them out across all task categories.}
\label{fig:PTvsSFT}
\end{figure}

\begin{figure}
\includegraphics[scale=0.55]{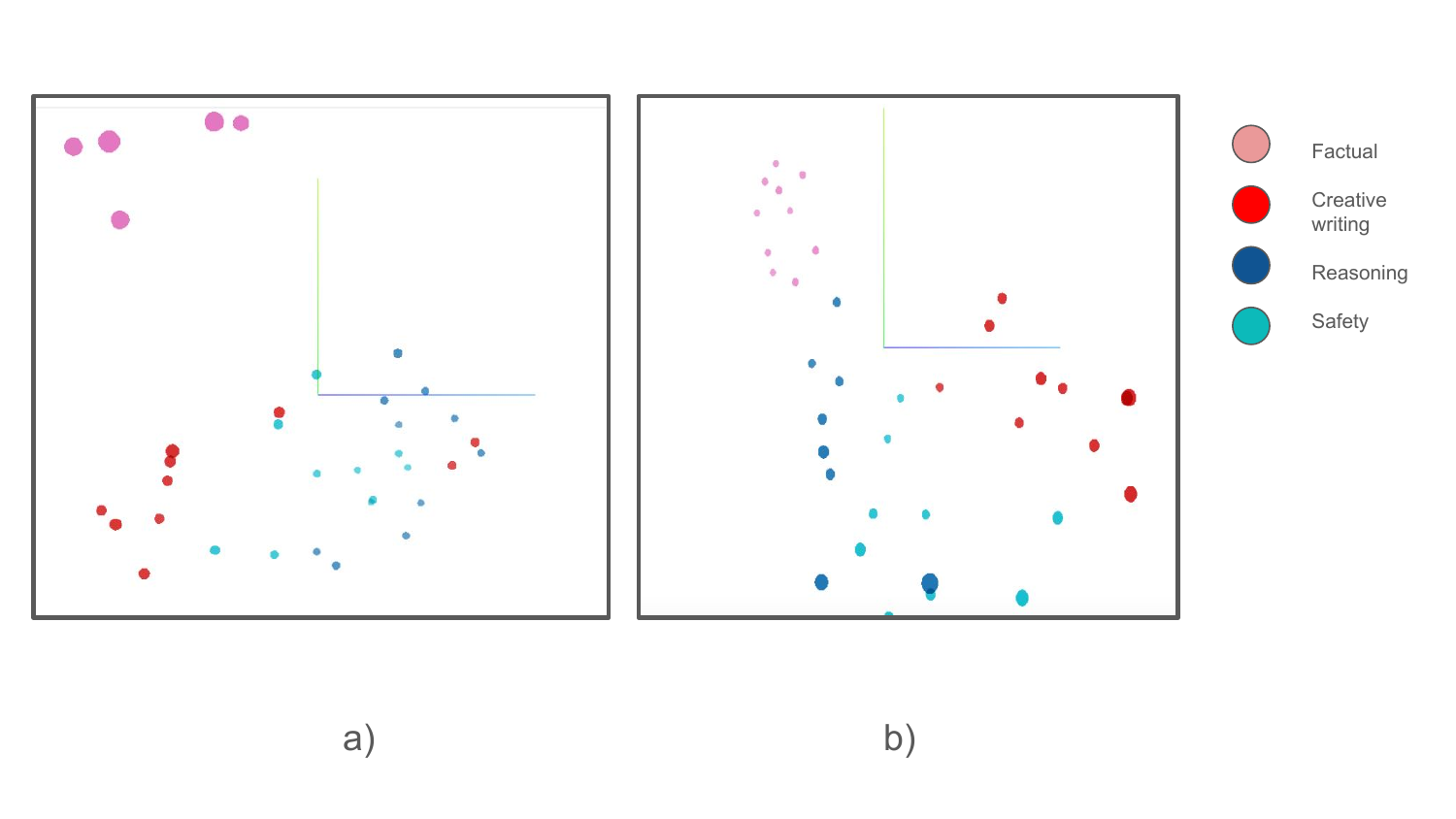}
\caption{(a) The UMAP plots for Llama2 7B SFT Chat model and (b) The UMAP plots for Krutrim model. In this plot, we compare the embedding projection for Llama2 7B SFT model against Krutrim model and show that Krutrim has better separability of tasks than Llama2 7B SFT model across all tasks while substantially surpassing in categories like ``creative writing".}
\label{fig:llamaVsKruSFT}
\end{figure}

We extract and analyse the embedding from the penultimate layer of Krutrim model using UMAP plots \cite{mcinnes2020umap} on data points which are randomly drawn from a myriad of task categories like factual, content writing, reasoning, safety etc. This further helps in analysing the impact of every stage of training like PT, SFT etc. as shown in figure \ref{fig:PTvsSFT}. UMAP, short for Uniform Manifold Approximation and Projection, serves as a dimension reduction method applicable not only for visualization akin to t-SNE but also for general non-linear dimension reduction purposes.
We also analyse LLaMA2 7B chat model\footnote{LLaMA 2-Chat 7B models are referred to as LLaMA 2 herein for brevity.} against Krutrim SFT model and show superiority overall while being substantially superior for creative generation task as shown in figure \ref{fig:llamaVsKruSFT} for English tasks. 

\subsection{Experiments on Probing Layer-Wise model learning}
\label{Sect:layerWise}
We analyze information present in layers at different depths by performing different probing tasks designed by chen et al.\cite{chen2023beyond}.
They perform analysis of LLaMA models of different sizes (7B, 13B, 70B parameters) tasked to answer complex multiple choice questions to capture models' performance on different aspects such as calculation, math problem solving (MPS), logical reasoning, truthfulness, and factual knowledge detection.
We perform similar evaluation of our language model to gain insights into the layer-level performance.
For a fair comparison, we perform the evaluation using the same test setting and the same datasets.
We use a partitions of GSM8K \cite{cobbe2021training} data for Math Problem Solving (MPS) Reasoning (MPS-Reason) and Calculation (MPS-Cal) tasks; Reclor \cite{yu2019reclor} for Logical Reasoning; and two tasks to capture two aspects of Hallucination Detection 1) Truthfulness using TruthfulQA MC task \cite{lin2022truthfulqa} and 2) Factual knowledge using LAMA \cite{petroni2019language}.
To analyze the Cross-Lingual capabilities of the model, we follow the xMPS-Reason task introduced by \cite{chen2023beyond} with a slight modification.  We replace all answers (as opposed to only the best answer) for every question from the MPS-Reason dataset with the machine translated answers, translated using IndicTrans2 \cite{gala2023indictrans2}.

\begin{figure*}[t!]
    \centering
    \begin{subfigure}[t]{0.5\textwidth}
        \centering
        \includegraphics[width=0.9\textwidth]{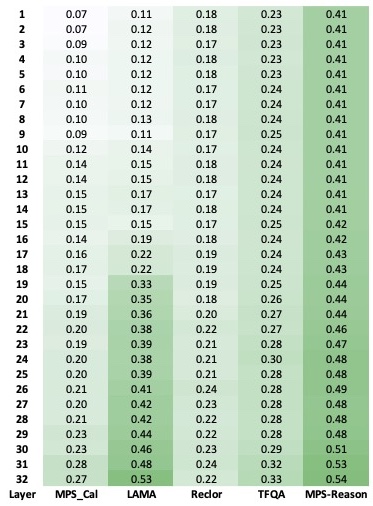}
        \caption{Accuracy of different layers across tasks}
        \label{fig:layerwiseTasks}
    \end{subfigure}%
    ~ 
    \begin{subfigure}[t]{0.64\textwidth}
        \centering
        \includegraphics[width=1.025\textwidth]{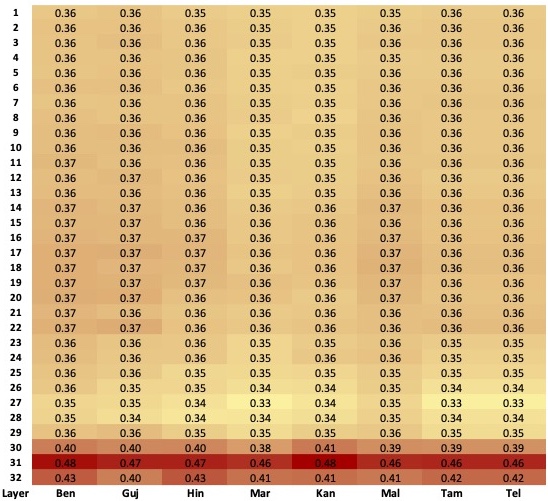}
        \caption{Accuracy on xMPS-Reason task across different cross-lingual en-INDIC Language pairs}
        \label{fig:layerlang}
    \end{subfigure}
    \caption{Heat-map depicting layer-wise performance of Krutrim LLM}
    \label{fig:layerwise}
\end{figure*}

We make the following observations from the experiments on English performed across different tasks on different layers of the model, as also depicted in Figure \ref{fig:layerwiseTasks}.
\begin{enumerate}
    \item \textbf{MPS-Cal:} Computational ability increases starting from a very low accuracy at the first layer to a significantly higher accuracy at the top layer.
    This observation aligns with that of LLaMA, making us speculate that the accuracy would further increase as we increase the depth of the model.
    \item \textbf{LAMA:} First few layers perform poorly on the task while the top few layers perform very well on the task.
    Our Model capture rich factual knowledge in it's top layers.
    \item \textbf{Reclor, TFQA, MPS-Reason:} A certain degree of the abstract knowledge and cognitive abilities are consistently present across all layers, although the best performance is observed in the final layers.
\end{enumerate}

We show the results of xMPS task aimed at cross-lingual analysis in Figure \ref{fig:layerlang}. 
We find that cross-lingual with all en-INDIC pairs follow a similar trend as English alone on the MPS-Reason task - a certain degree of the mathematical reasoning capabilities are consistently present across all layers, although the performance is a bit lower as compared to English-only.
We also notice a spike in the performance in the last 2-3 layers. 
Another observation in align with \cite{chen2023beyond} is that the penultimate layer outperforms the last layer in all xMPS experiments. 

\subsection{Standard Indic Benchmarks}

Traditional metrics like BLEU, ROUGE, and GLUE have long been the go-to measures for assessing the performance of language generation systems. However, these metrics often fall short when it comes to capturing the nuanced semantic similarity between sentences \cite{Peng_2017}. In the table \ref{tab:scores} summary, we examine the evaluation results of the sentences \textit{"Here is the footwear"} and \textit{"Here are the shoes"} using BLEU, ROUGE, and GLUE scores. Despite the apparent similarity in meaning between these sentences, the traditional metrics yield scores that do not accurately reflect their semantic equivalence. This highlights the limitations of conventional evaluation methods and underscores the need for more sophisticated approaches, such as BERT score, which can provide a deeper understanding of contextual semantics in language generation tasks.

\begin{table}[htbp]
  \centering
  \caption{Summary on study of "Evaluation metric and Scores"}
  \label{tab:scores}
  \begin{tabular}{@{}ll@{}}
    \toprule
    \textbf{Metric} & \textbf{Score} \\
    \midrule
    BLEU & $\{ \text{precisions}: [0.5, 0.0, 0.0] \}$ for unigram, bigram and trigram matches. \\
    ROUGE & $\{ \text{rouge1}: 0.5, \text{rouge2}: 0.0, \text{rougeL}: 0.5, \text{rougeLsum}: 0.5 \}$ \\
    GLUE & $\{ \text{pearson}: 0.7287, \text{spearmanr}: 0.6787 \}$ \\
    BERT score & \{0.86\} \\
    \bottomrule
  \end{tabular}
\end{table}

Hence, we use metrics for evaluating Krutrim across mentioned tasks in table \ref{tab:metrics}. We used subset of IndicXtreme benchmark tasks \cite{Doddapaneni2022towards} for evaluation against other standard LLMs which have indic generation capabilities.

\begin{table}[htbp]
  \centering
  \begin{tabular}{@{}ll@{}}
    \toprule
    \textbf{Task type} & \textbf{Metric} \\
    \midrule
    Translation & BERTScore \\
    Summarisation & BERTScore \\
    Paraphrasing & BERTScore \\
    Reading comprehension & BERTScore \\
    Classification (e.g., sentiment, MCQs) & Accuracy \\
    \bottomrule
  \end{tabular}
\caption{Evaluation Metrics for Various Tasks}
  \label{tab:metrics}
\end{table}

\subsubsection{COPA}

We assess the IndicCOPA dataset, utilizing a three-shot setting for evaluation, akin to practices in language modeling. By translating prompts and keywords, we ensure consistency across languages. Krutrim LLM demonstrates superior performance across languages, except Sanskrit. Notably, in Malayalam and Telugu, it closely aligns with GPT-3.5's performance. Refer to Table \ref{IndicCOPA_tab} for a comparative analysis against other models across 10 languages.

\begin{table}[h]
\centering

\begin{tabular}{ccccccccc}
    \toprule
\textbf{Model} &  \textbf{bn} & \textbf{gu} & \textbf{hi} & \textbf{kn} & \textbf{ml} & \textbf{mr} & \textbf{ta} & \textbf{te}  \\
\midrule
\textbf{Krutrim LLM} & \textbf{0.89} & \textbf{0.83} & \textbf{0.86} & \textbf{0.88} & \textbf{0.88} & \textbf{0.87} & \textbf{0.89} & \textbf{0.89}  \\
\textbf{GPT-3.5} & 0.77 & 0.73 & 0.77 & 0.74 & 0.75 & 0.7 & 0.72 & 0.75 \\
\textbf{Airavata} & - & - & 0.74 & -  & -  & -  & -  & -  \\ 
\textbf{KAN-LLaMA} & - & -  & -  & 0.74  & -  & -  & -  & -    \\
\textbf{TAM-LLaMA}  & -  & -  & -  & -  & -  & -  & 0.77  & -  \\
\bottomrule
\end{tabular}
\captionsetup{justification=centering, skip=10pt}
\caption{Comparison of GPT-3.5, Airavata, Tamil-LLaMA, Kannada-LLaMA with our model, on IndicCOPA dataset (BERTScore), same cleaning methods have been applied to all model results.}
\label{IndicCOPA_tab}

\end{table}

\subsubsection{QA}

Our evaluation extends to the IndicQA dataset sourced from the AI4Bharat IndicXtreme benchmark. This zero-shot assessment involves questions derived from provided contexts, with prompt translations ensuring language parity. Employing semantic matching metrics, our model exhibits an 85\% superiority over Tam-LLaMA. See Table \ref{IndicQA_tab} for a comprehensive comparison across 9 languages.

\begin{table}[h]
\centering
\begin{tabular}{cccccccccc}
\toprule
\textbf{Model} &  \textbf{bn} & \textbf{gu} & \textbf{hi} & \textbf{kn} & \textbf{ml} & \textbf{mr} & \textbf{ta} & \textbf{te} \\
\midrule
\textbf{Krutrim LLM}  & \textbf{0.65} &\textbf{0.64} & \textbf{0.64} & \textbf{0.60} & \textbf{0.66} & \textbf{0.58} & \textbf{0.75} & \textbf{0.83}  \\
\textbf{Airavata} & - & - & 0.62 & -  & -  & -  & -  & -   \\
\textbf{KAN-LLaMA} & - & -  & -  & 0.52  & -  & -  & -  & -    \\
\textbf{TAM-LLaMA}  & -  & -  & -  & -  & -  & -  & 0.35  & -  \\
\bottomrule
\end{tabular}
\captionsetup{justification=centering, skip=10pt}
\caption{Comparison of Airavata, Tamil-LLaMA, Kannada-LLaMA with our model on IndicQA dataset (BERTScore), same cleaning methods have been applied to all model results.}
\label{IndicQA_tab}
\end{table}

\subsubsection{Sentiment}

Delving into the IndicSentiment dataset \cite{Doddapaneni2022towards}, our model showcases remarkable performance differentials across languages. Particularly in Tamil and Telugu, it surpasses GPT-3.5 by more than 0.8 points. Utilizing accuracy metrics and a three-shot evaluation approach, our model's superiority is evident across various languages. The comparative results are detailed in Table \ref{IndicSentiment_tab}.

\begin{table}[h]

\centering
\begin{tabular}{cccccccccc}
\toprule
\textbf{Model} &  \textbf{bn} & \textbf{gu} & \textbf{hi} & \textbf{kn} & \textbf{ml} & \textbf{mr} & \textbf{ta} & \textbf{te} \\
\midrule
\textbf{Krutrim LLM} & \textbf{0.95} & \textbf{0.96} & \textbf{0.96} & \textbf{0.95} & \textbf{0.96} & \textbf{0.97} & \textbf{0.94} & \textbf{0.95}  \\
\textbf{GPT-3.5} & 0.50 & 0.81 & \textbf{0.96} & 0.60 & 0.75 & 0.88 & 0.51 & 0.53 \\
\textbf{Airavata} & - & - & 0.84 & -  & -  & -  & -  & -   \\
\textbf{KAN-LLaMA} & - & -  & -  & 0.85  & -  & -  & -  & -    \\
\textbf{TAM-LLaMA}  & -  & -  & -  & -  & -  & -  & 0.78  & -  \\
\bottomrule
\end{tabular}
\captionsetup{skip=10pt} 
\caption{Comparison of GPT-3.5, Airavata, Tamil-LLaMA, Kannada-LLaMA with our model on IndicSentiment (Accuracy) dataset, same cleaning methods have been applied to all model results.}
\label{IndicSentiment_tab}

\end{table}


\subsubsection{Translation}

Our examination extends to the IndicTranslation dataset sourced from AI4Bharat IndicXtreme. Adopting a three-shot evaluation approach, our model outperforms GPT-3.5 in Gujarati, Kannada, Malayalam, and Telugu by substantial margins. Comparative analysis across 7 languages is presented in Table \ref{IndicTranslation_tab}.

\begin{table}[h]
\centering

\begin{tabular}{cccccccccc}
\toprule
\textbf{Model}  &  \textbf{bn} & \textbf{gu} & \textbf{hi} & \textbf{kn} & \textbf{ml} & \textbf{mr}  & \textbf{te} \\
\midrule
\textbf{Krutrim LLM} & \textbf{0.88} & \textbf{0.89} & \textbf{0.95} & \textbf{0.88} & \textbf{0.89} & \textbf{0.92} & \textbf{0.88}    \\
\textbf{GPT-3.5} & 0.72 & 0.54 & 0.78 & 0.54 & 0.57 & 0.71 & 0.34 \\
\textbf{Airavata} & - & - & 0.94 & -  & -  & -   & -    \\
\textbf{KAN-LLaMA} & - & -  & -  & 0.59  & -  & - & -     \\
\bottomrule
\end{tabular}
\captionsetup{skip=10pt} 
\caption{Comparison of GPT-3.5, Airavata, Tamil-LLaMA, Kannada-LLaMA with our model on IndicTranslation dataset (BERTScore). Same cleaning methods have been applied to all model results.}
\label{IndicTranslation_tab}

\end{table}

\subsubsection{Paraphrase}

The evaluation encompasses the IndicXParaphrase dataset, where pairs of sentences are classified as supportive (1) or not (0). Our model demonstrates superior performance across languages, except Bengali. Detailed results are tabulated in Table \ref{IndicXParaphrase_tab}.

\begin{table}[h]
\centering
\begin{tabular}{cccccccccc}
\toprule
\textbf{Model} &  \textbf{bn} & \textbf{gu} & \textbf{hi} & \textbf{kn} & \textbf{ml} & \textbf{mr} & \textbf{te} \\
\midrule
\textbf{Krutrim LLM}& \textbf{0.91} & - & \textbf{0.97} & \textbf{0.82} & \textbf{0.90} & \textbf{0.94}  & \textbf{0.61}   \\
\textbf{GPT3.5} & 0.72 & \textbf{0.54} & 0.78 & 0.54 & 0.57 & 0.71 & 0.34 \\
\textbf{Airavata} & - & - & 0.60 & -  & -  & -   & -   \\
\textbf{KAN-LLaMA} & - & -  & -  & 0.59  & -  & -    & -    \\
\bottomrule
\end{tabular}
\captionsetup{skip=10pt} 
\caption{Comparison of GPT-3.5, Airavata, Tamil-LLaMA, Kannada-LLaMA with our model on IndicXParaphrase (Accuracy), same cleaning methods have been applied to all model results.}
\label{IndicXParaphrase_tab}
\end{table}

\subsection{Standard English Benchmarks}
\begin{figure}
\centering
\includegraphics[width=0.6\textwidth]{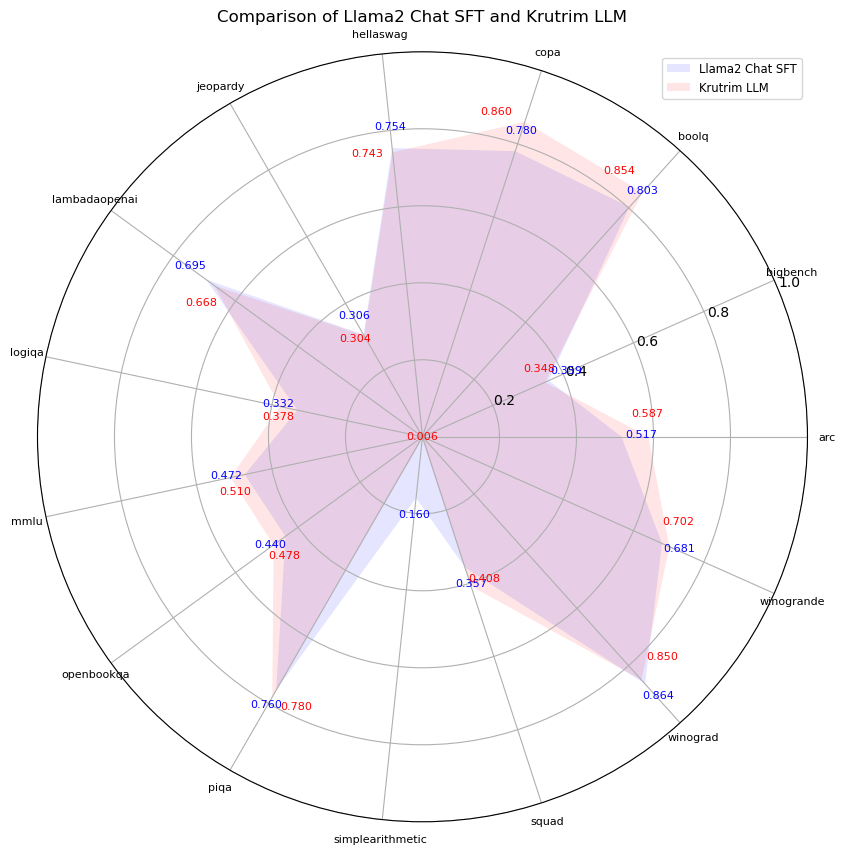}
\caption{Radar chart for comparison of LLaMA chat model with Krutrim model on English benchmarks.}
\label{fig:radar}
\end{figure}

Recent advancements such as LLAMA 3 \cite{llama3} and Mistral 7B \cite{jiang2023mistral} have demonstrated impressive capabilities. However, these models are trained with significantly higher computational resources and larger datasets compared to Krutrim model. Additionally, they do not perform well on Indic languages. Consequently, we have not included comparisons with these models due to these substantial differences in FLOPS.
We evaluate Krutrim fine tuned model on various English benchmark tasks and compare against Llama-2 7B chat SFT. The results are listed in table \ref{llamaVsKrutrim}, where we surpass Llama on 10 out of 17 tasks. The graph in figure \ref{fig:radar} also elucidates the same. 

\begin{table}
\centering
\begin{adjustbox}{center}
\centering
\resizebox{9cm}{!}
{
\begin{tabular}{ccc}
\toprule

\textbf{Task} &\textbf{Llama2 Chat SFT} &\textbf{Krutrim LLM} \\
\midrule
ARC &0.517 & \textbf{0.587}  \\
BIG-Bench &\textbf{0.359} & 0.348  \\
BoolQ &0.803 & \textbf{0.854} \\
COPA &0.78 & \textbf{0.86}  \\
Hella Swag &\textbf{0.754} & 0.743 \\
Jeopardy &\textbf{0.306} & 0.304  \\
LAMBADA OpenAI &\textbf{0.695} & 0.668  \\
LogiQA &0.332 & \textbf{0.378}  \\
MathQA &0.436 & \textbf{0.440}\\
MMLU &0.472 & \textbf{0.51}  \\
Openbook QA &0.44 & \textbf{0.478}  \\
PIQA &0.760 & \textbf{0.78}  \\
Simple Arithmetic &\textbf{0.160} & 0.06 \\
Squad &0.357 & \textbf{0.408} \\
Winograd &\textbf{0.864} & 0.85  \\
Winogrande &0.681 & \textbf{0.702}  \\
\midrule
\textbf{Average} &\textbf{0.552} & \textbf{0.569} \\
\bottomrule
\end{tabular}
}
\end{adjustbox}
\captionsetup{justification=centering, skip=10pt}
\caption{Comparison of LLaMA chat model with Krutrim model.}
\label{llamaVsKrutrim}
\end{table}

\FloatBarrier

\FloatBarrier
\subsection{Human Evaluations}

\begin{figure}[h]
  \centering
  \includegraphics[width=1\textwidth]{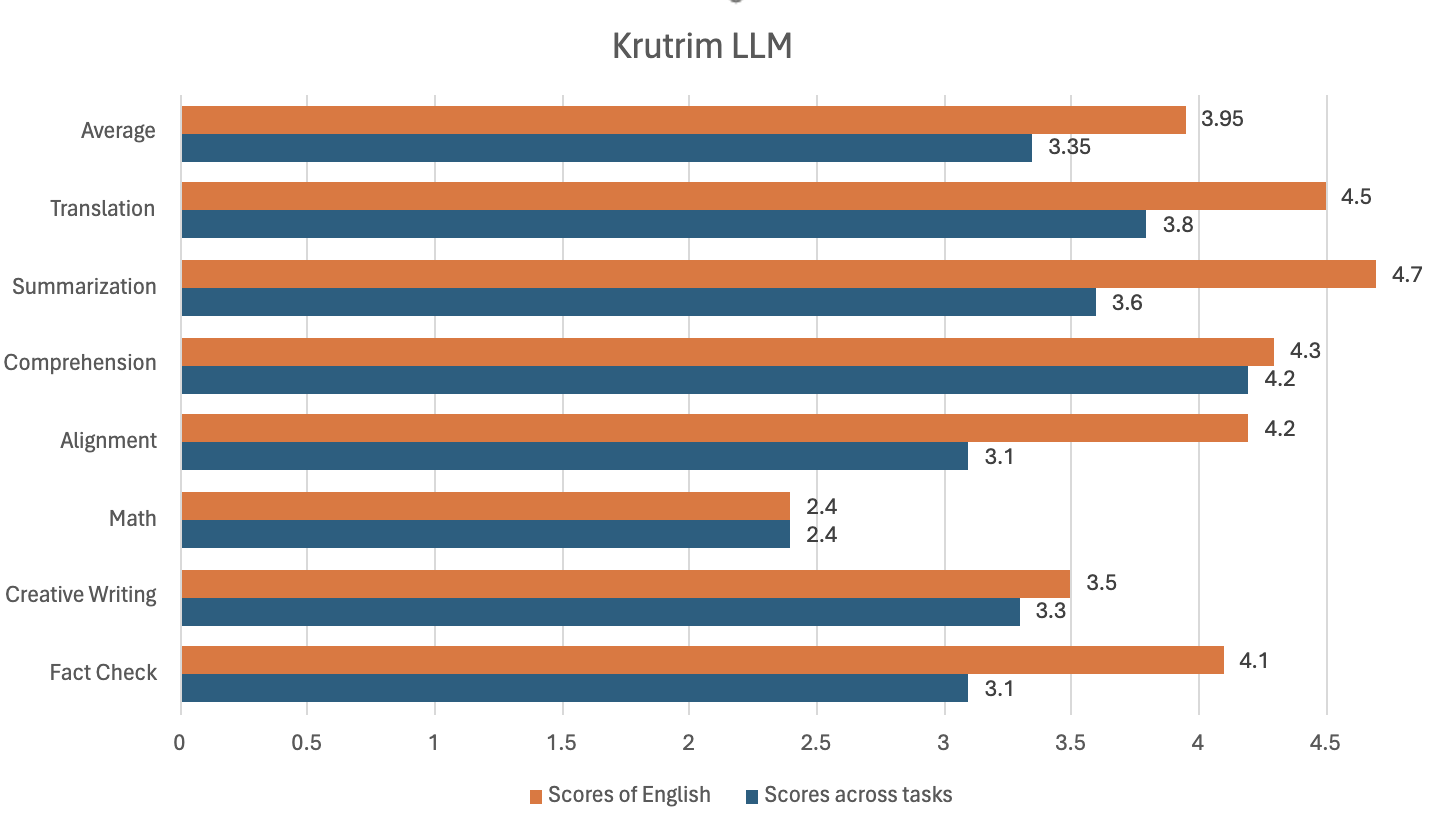} 
  \caption{Comparison of scores across languages in various categories of tasks.}
  \label{fig:example}
\end{figure}

\FloatBarrier

\FloatBarrier
\subsubsection{
Continual Pre-training: Techniques and Result}
\begin{figure}
\centering
\includegraphics[scale=0.4]{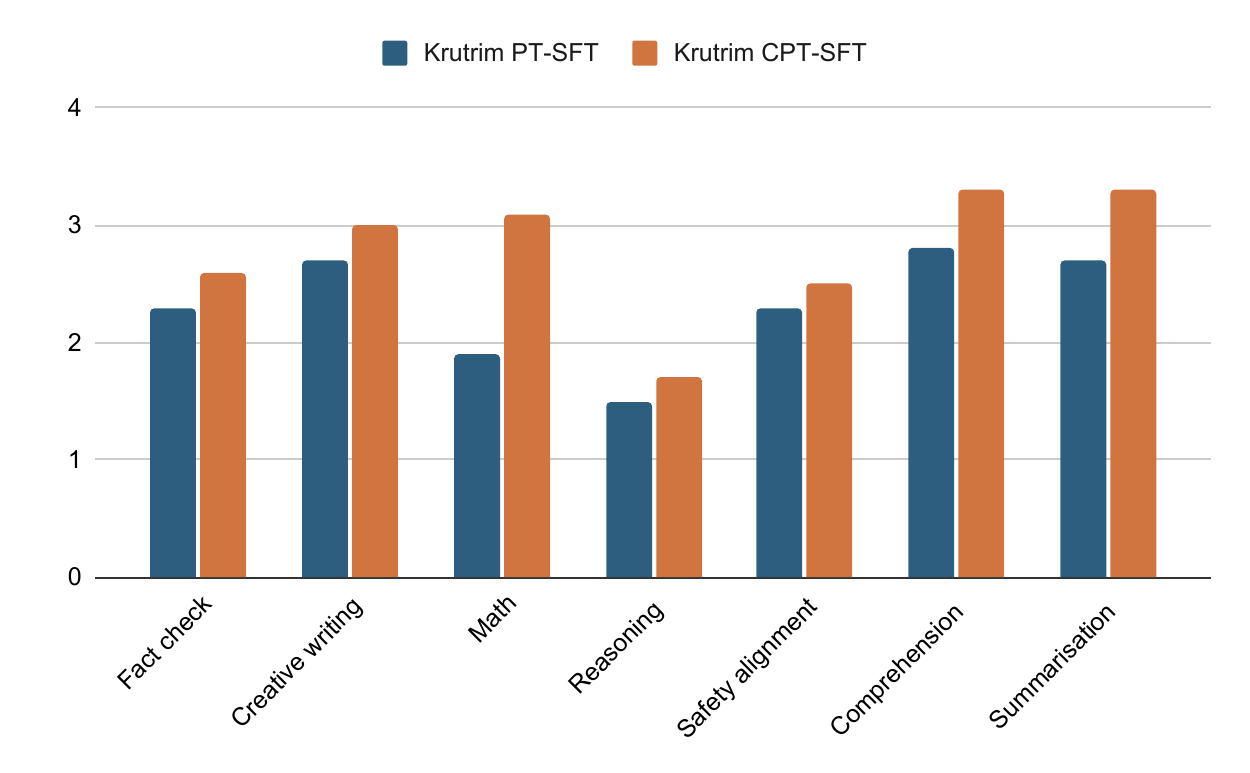}
\caption{Qualitative comparison after pre-training and continual pre-training. This plot shows the gain across languages and various tasks categories after performing Continual Pre-training CPT.}
\label{PT-SFTvsCPT-SFT}
\end{figure}

The CPT data should include some of the data from the original pre-training mix. A decent starting point is 25\% original data and 75\% new data. The new data ought have better quality. These proportions can be adjusted based on the amount of forgetting we're seeing. The learning rate should be adjusted to pick from the last value where it was stopped in the pre-training step.\\
To emphasize the impact of CPT, we present the gain in the MOS (Mean Opinion Score) of qualitative evaluation performed across 7 task categories averaged for major Indic languages in chart \ref{PT-SFTvsCPT-SFT}.  We perform two experiments to obtain two models "Krutrim PT-SFT" and "Krutrim CPT-SFT" respectively. In first experiment, we perform small scale SFT on pre-trained model (PT) to obtain "Krutrim PT-SFT". While in second experiment, "Krutrim CPT-SFT" is obtained by performing SFT with same data as was used in experiment 1 but on continually pre-trained model (CPT). We find that SFT on CPT model substantially outperforms SFT on PT model across tasks qualitatively.

\FloatBarrier

\subsection{SFT for Answering Factual Questions}
To analyze the impact of the experiment mentioned in Subsection \ref{webrag}, we conducted a qualitative study. In this study, we analyzed a subset of real-world factual questions posed by users. The testing set was organized into three distinct categories.
\begin{enumerate}
    \item Factual Questions
    \item Ambiguous or Factually Incorrect Questions
    \item Generation-Based Questions
\end{enumerate}
We employed a data tagging team to manually assess the responses generated by Krutrim model. These assessments were categorized into the following scenarios with corresponding labels:
\begin{itemize}
    \item Good: The response accurately addresses the query.
    \item Bad: The response fails to address the query correctly.
    \item Refrained: The model deliberately abstains from answering or does not provide relevant information.
\end{itemize}

\begin{figure}
\centering
\includegraphics[scale=0.43]{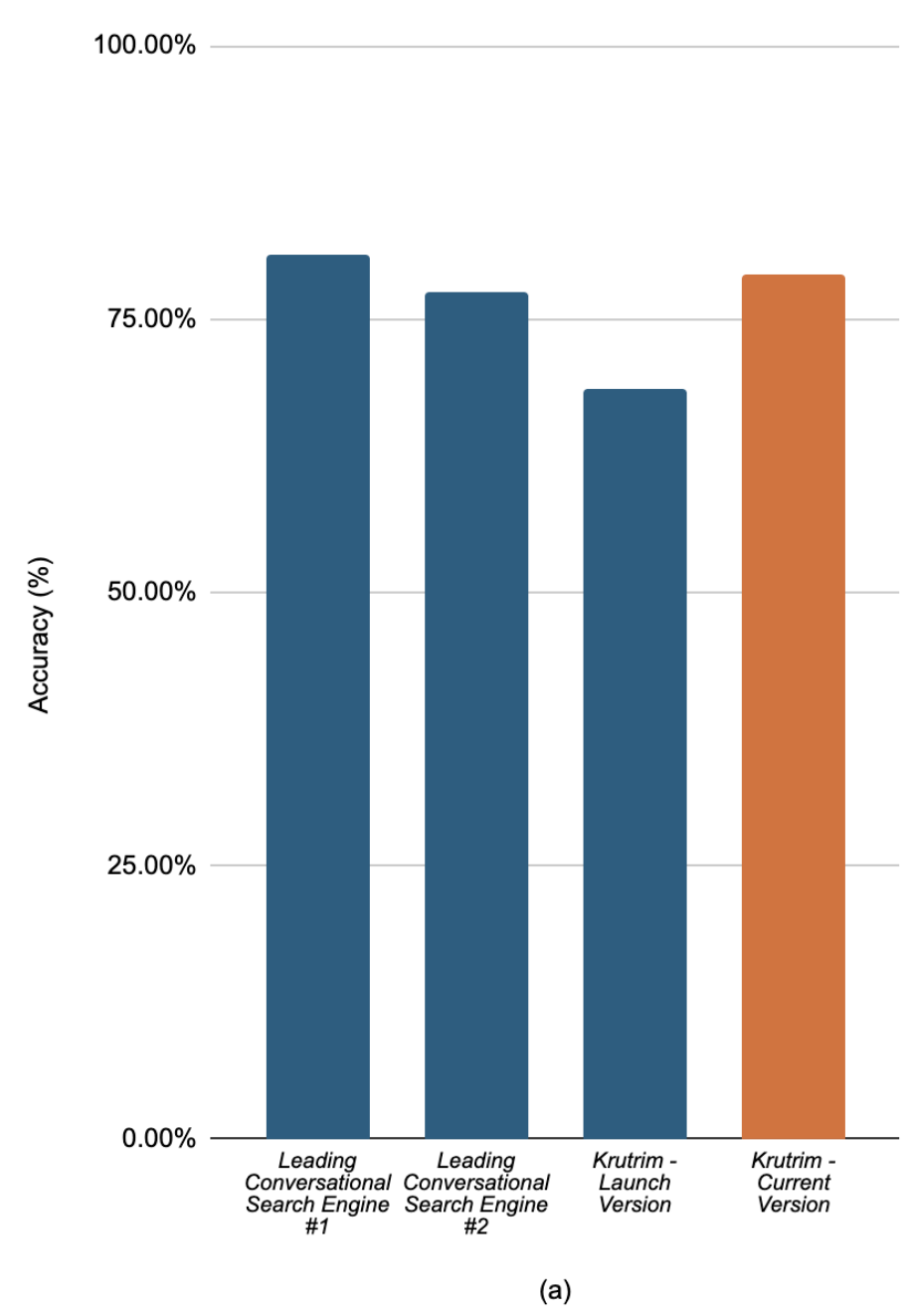}
\includegraphics[scale=0.43]{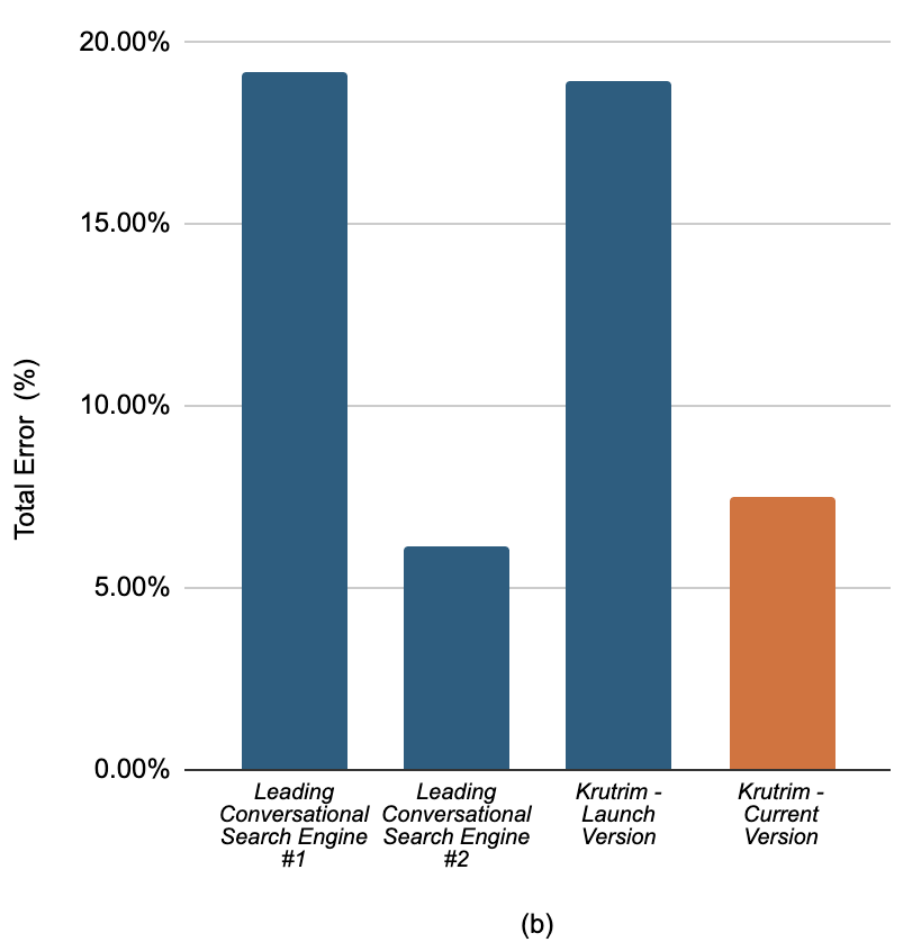}
\caption{Performance comparison of Krutrim with Leading Conversational Search Engine(s) on the task on  Factual Question Answering. The plot on the left shows the (\%)age accuracy (higher the better) of the three models. The plot on the right shows the (\%)age error rate (lower the better) for the three models.}
\label{webrag-performance}
\end{figure}

We present a comparative analysis of accuracy and error rates across different versions of Krutrim and two leading conversational search engines on our test set. The results are detailed as follows:

Accuracy Percentages on the Test Set:
\begin{itemize}
\item Leading Conversational Search Engine \#1: Achieved an accuracy of 80.87\%.
\item Leading Conversational Search Engine \#2: Recorded an accuracy of 77.43\%.
\item Krutrim - Launch Version: Initially posted an accuracy of 68.67\%.
\item Krutrim - Current Version: Demonstrated a significant improvement, reaching an accuracy of 79.13\%.
\end{itemize}
Error Percentages on the Test Set:
\begin{itemize}
\item Leading Conversational Search Engine \#1: Registered an error rate of 19.17\%.
\item Leading Conversational Search Engine \#2: Showed a lower error rate of 6.13\%.
\item Krutrim - Launch Version: Had an error rate of 18.93\%.
\item Krutrim - Current Version: Marked improvement is seen with a reduced error rate of 7.47\%.
\end{itemize}
These findings are visually represented in Figures \ref{webrag-performance}.b and \ref{webrag-performance}.b, providing a clear depiction of the performance metrics across the different systems.

\FloatBarrier

\section{Qualitative Analysis}

\begin{figure}

\frame{\includegraphics[width=1\textwidth]{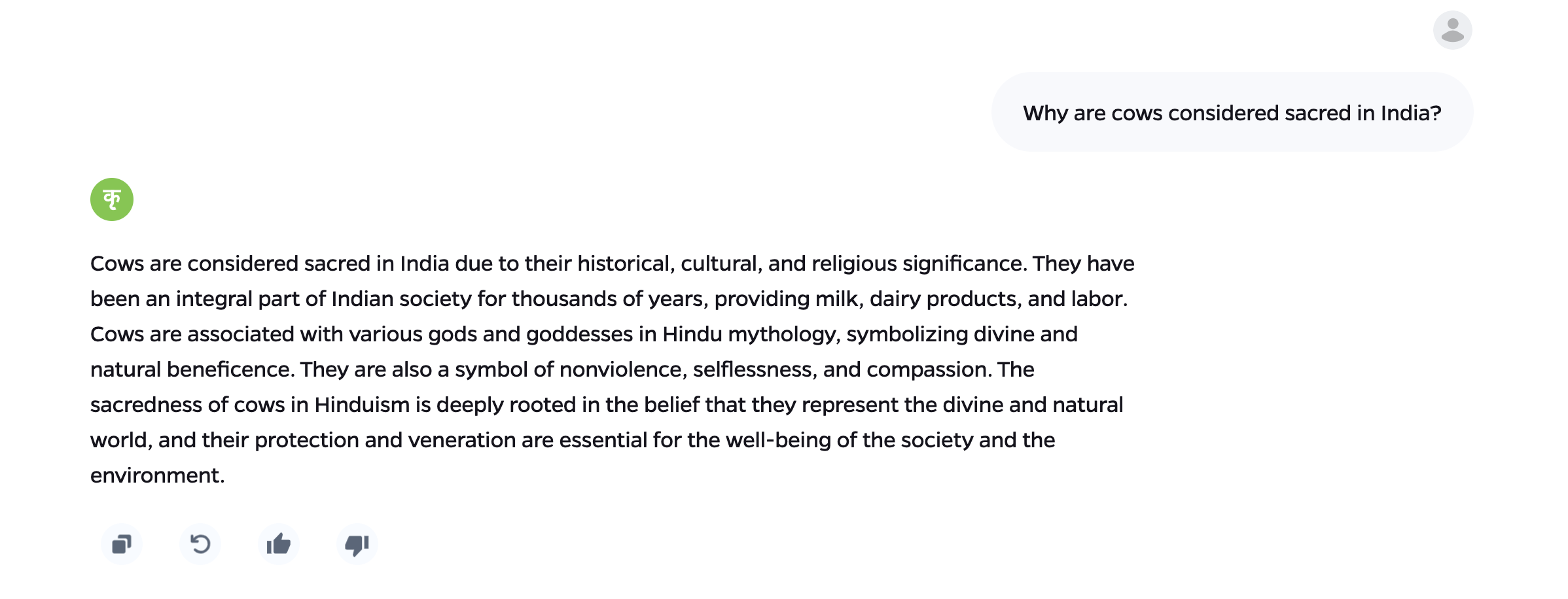}}
\caption{Cultural Sensitivity: Krutrim's ability to respect cultural practices. Krutrim is unbiased and does not favour one religion or caste.}
\end{figure}

\begin{figure}

\frame{\includegraphics[width=1\textwidth]{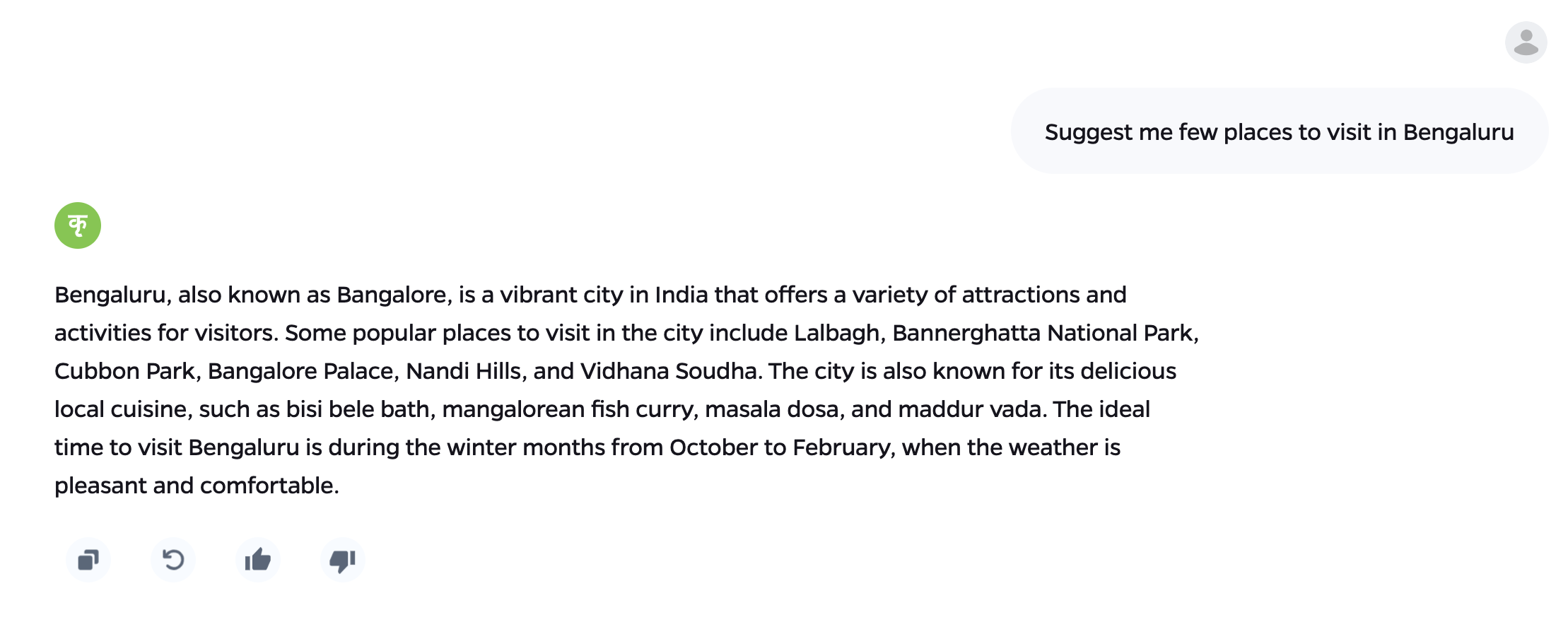}}
\caption{Local Relevance: Krutrim can also recommend places, restaurants locally.}
\end{figure}

\begin{figure}

\frame{\includegraphics[width=1\textwidth]{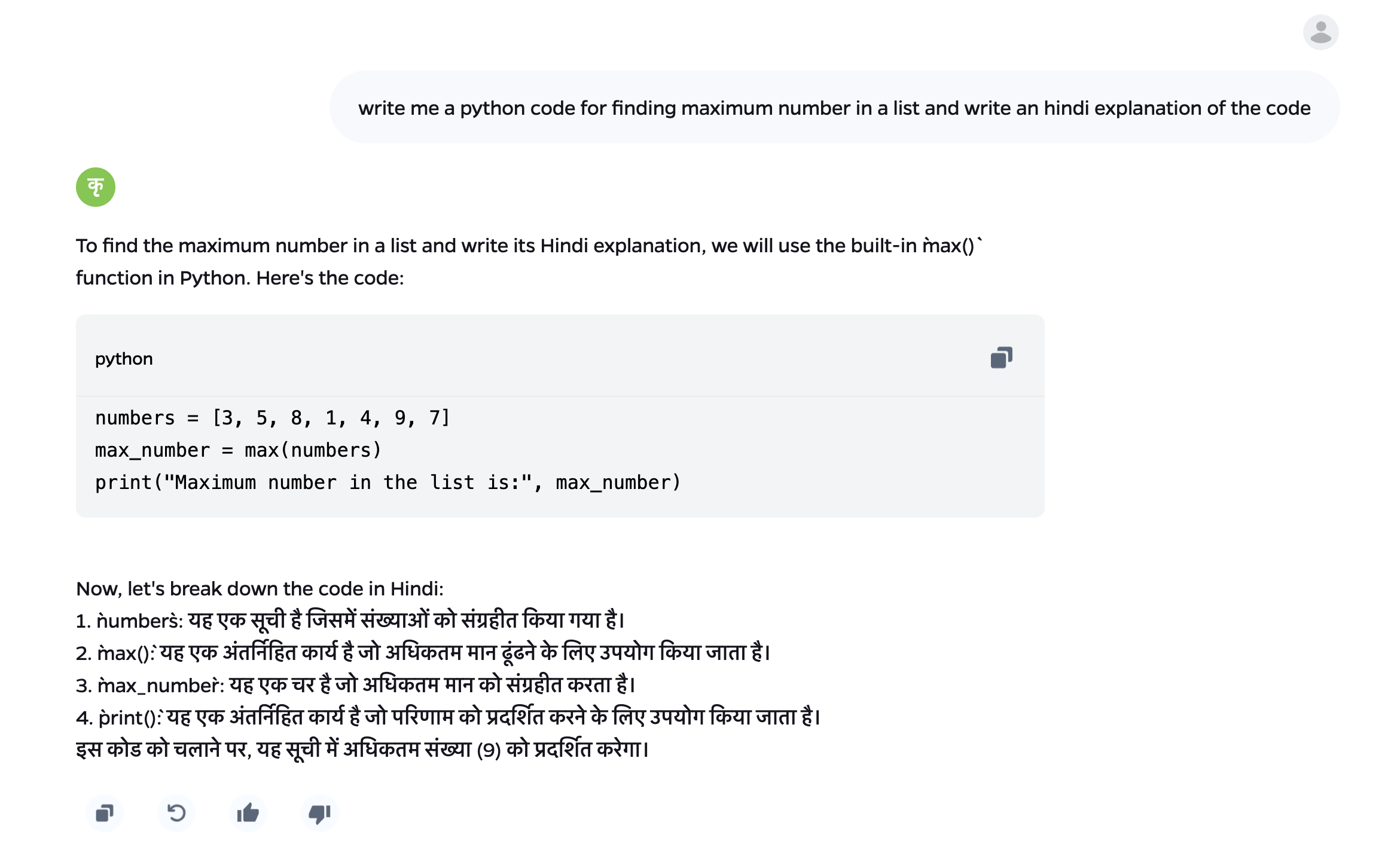}}
\caption{Coding abilities: This shows the coding ability of Krutrim and explaining the code in Indic languages}
\end{figure}

\begin{figure}

\frame{\includegraphics[width=1\textwidth]{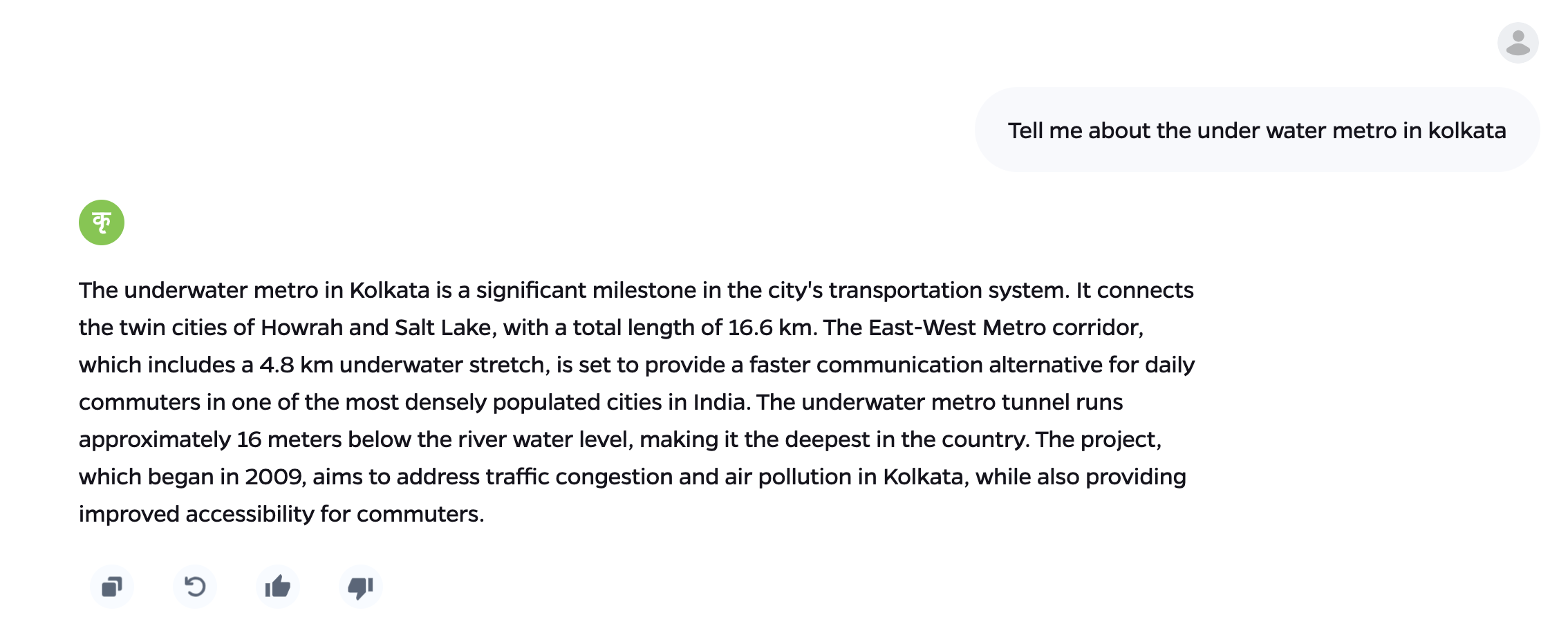}}
\caption{News and latest trends: Krutrim can answer the latest news trends as it also has robust WebRAG pipeline for factual queries.}
\end{figure}

\begin{figure}

\frame{\includegraphics[width=1\textwidth]{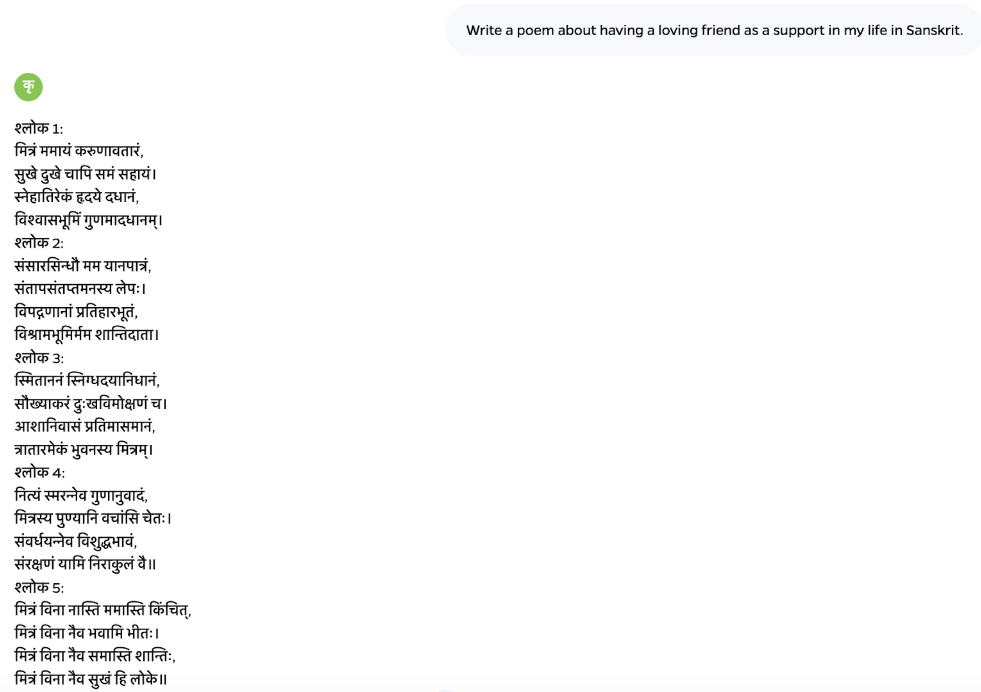}}
\caption{Sanskrit generation: Krutrim can do creative writing in Sanskrit}
\end{figure}

\begin{figure}
\centering
\includegraphics[width=1\textwidth]{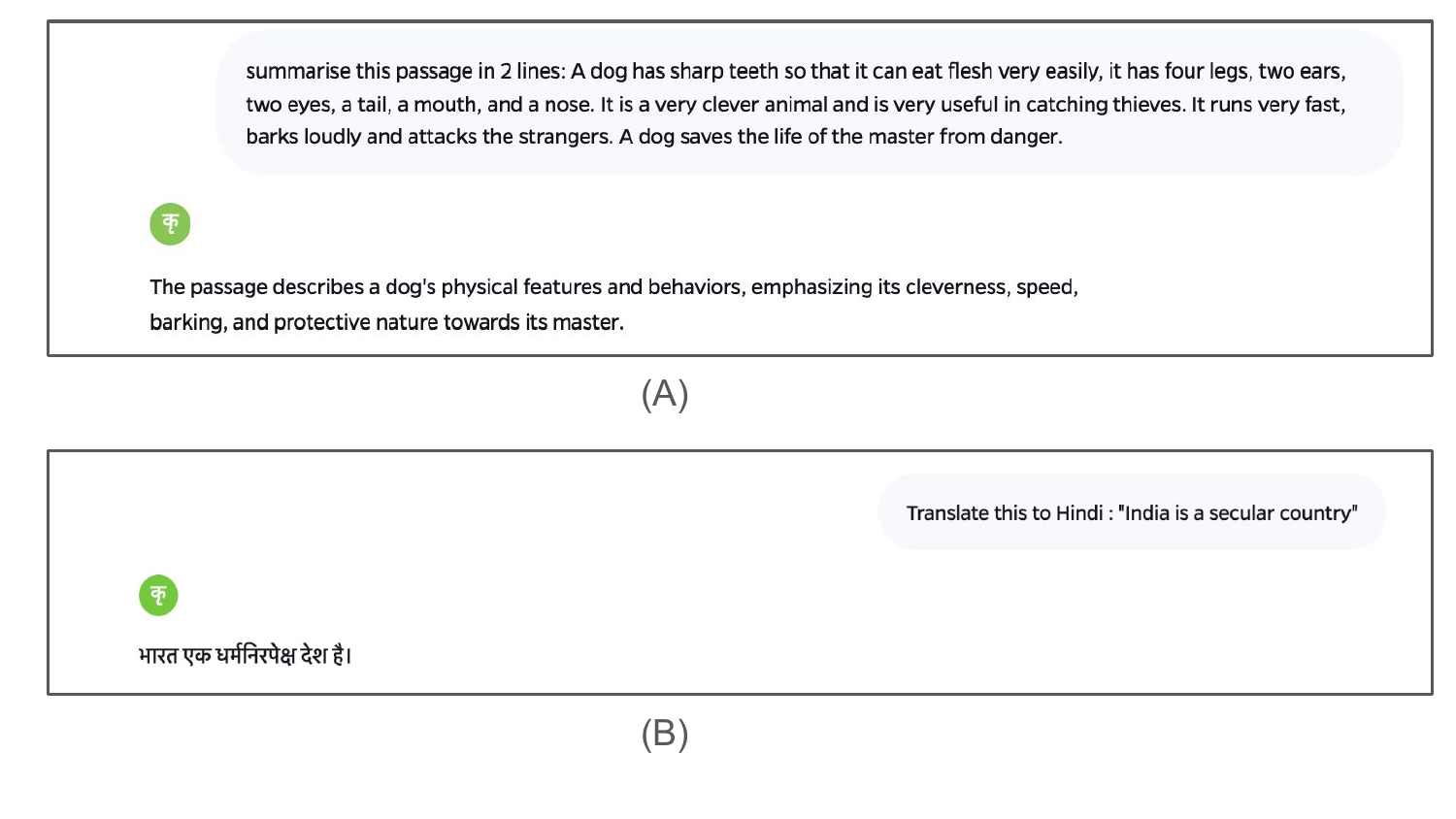}
\caption{ Krutrim is able to perform on various tasks across  Languages. It demonstrates proficiency in Summarisation and Translation tasks on Indic languages as shown in (A) and  (B)}
\end{figure}

\FloatBarrier

\section{Conclusion}
Krutrim establishes itself as a pioneering effort in developing the first large language model (LLM) specifically catering to the Indian context. This model addresses the crucial aspects of encompassing cultural intricacies and rich linguistic landscape, incorporating a vast amount of Indic data, and ensuring accessibility for the extensive Indian population.

Krutrim stands out by leveraging the largest known collection of Indian data for training, achieving performance that matches or sometimes surpasses several contemporary LLMs in India context, even those boasting significantly larger parameter sizes for general applications.  Furthermore, Krutrim integrates real-time web search capabilities, empowering it to deliver factual and up-to-date information outperforming state of the art system such on adversarial and factually incorrect questions.

To facilitate user interaction, Krutrim LLM is readily available through a user-friendly conversational interface accessible at \url{https://chat.olakrutrim.com}. This accessibility aspect positions Krutrim to broadly contribute across various sectors in India and serve as a foundation for further advancements in Indian-centric AI applications. 

\section{Authorship, Recognition of Contributions, and Acknowledgements
}
Please cite this work as ``Krutrim (2024)”.

\textbf{Pre-training}:
Aditya Kallappa, Palash Kamble, Abhinav Ravi, Akshat Patidar, Deepak Kumar, Vinayak Dhruv,  
 Raghav Awasthi, Arveti Manjunath, Gautam Bhargava, Chandra Khatri
\\~\\
\textbf{Fine-tuning and Alignment}
Aditya Kallappa, Palash Kamble, Abhinav Ravi,  Kumar Ashish,
Deepak Kumar, Sanket Shah, Sulabh Katiyar,  Vinayak Dhruv, Sindhu Pawar, Soham Pendurkar
\\~\\
\textbf{Evaluations} Arveti Manjunath, Pranav Raveendran, Bidyapathi Ray
\\~\\
\textbf{Data and Tokenisation}:
Aditya Kallappa, Palash Kamble, Akshat Patidar, Vinayak Dhruv, Daud Ibrahim, Divyansh Rajput, Pidathala Sowjanya, Rahul Kumar, Rishabh Nahata, Pranav Raveendran, Bidyapathi Ray
\\~\\
\textbf{Deployment}:
Prateek Shrivastava, Ashok Jagannathan, Raghav Awasthi, Yogendra Mishra 
\\~\\
\textbf{Web experience}:
Azhagiri S, Priyanka Nayak, Sandesh Bafna
\\~\\
\textbf{Additional contributions}:
Aniruddha Uttam Tammewar, Vivek Dahiya, Ali Faraz, Ayush Tarun, Shaharukh Khan, Debanjana Biswas, Shubham Agarwal, Ashish Anand Kulkarni, Rajkiran Panuganti, Hareesh Kumar, Shubham Kakde, Nishant kumar, Himanshu Gupta
\\~\\
\textbf{Acknowledgement}: Ravi Jain, Bhavish Aggarwal


\nocite{*} 

\printbibliography

\medskip

{
\small
}

\end{document}